\documentclass{article}

\usepackage[preprint]{neurips_2025}
\makeatletter
\renewcommand{\@neuripsyear}{2026}
\renewcommand{\@neuripsordinal}{40th}
\makeatother

\usepackage[utf8]{inputenc}
\usepackage[T1]{fontenc}
\usepackage{hyperref}
\usepackage{url}
\usepackage{booktabs}
\usepackage{array}
\usepackage{amsfonts}
\usepackage{nicefrac}
\usepackage{microtype}
\usepackage{xcolor}
\usepackage{graphicx}
\usepackage{amsmath}
\setcitestyle{round}
\hypersetup{hidelinks}

\title{Proxy Confidence: Auditing Black-Box LLM Agents with a Surrogate's Log-Probabilities}
\workshoptitle{Interpreting Agent Behavior (IAB)}

\author{%
  Yikai Zhao\\
  Amazon\\
  \texttt{yikai@amazon.com}\\
  \And
  Saurabh Pandey\\
  Amazon\\
  \texttt{saurabpd@amazon.com}\\
  \And
  Pradeep Kumar Misra\\
  Amazon\\
  \texttt{pkmisra@amazon.com}\\
}

\begin{document}
\maketitle

\begin{abstract}
A deployed LLM agent emits tool calls, queries, and code that can be silently wrong---and by the time the
error surfaces, the action has already run. The frontier chat APIs these agents run on offer no reliable way
to tell which calls to trust. They hide the model's own token probabilities. The agent's stated confidence
barely beats chance on the mistakes that matter. And resampling does not help either, since frontier
models are now known to be highly repetitive, reproducing the same call across samples.

We recover the missing signal from a low-cost open-weight surrogate run in parallel. It reads the
same context, schema, and proposed action as the agent, then scores the call from its \emph{own}
log-probabilities through a family of complementary readouts. \emph{Teacher forcing} and \emph{request-PMI}
(pointwise mutual information) weigh the likelihood of each argument value; a \emph{discriminative verdict}
judges the call as a whole; tool-choice competition tests the function against its siblings. One principle
tells an operator which to trust: a generative likelihood localizes wrong argument \emph{values}, while the
verdict catches holistically wrong calls. When the error type is unknown, an ensemble over the family is the
low-regret default---a light logistic combiner, matched by a simple orientation-corrected average. The whole
readout is training-free, needs no access to the agent's internals, and costs one prefill pass in parallel
with the tool call.

This proxy ranks correct calls above wrong ones more reliably than either signal a black-box operator
has today. Its edge over an agent's stated confidence is largest exactly where it matters most: on
confident mistakes. On difficult coding tasks it reaches AUROC $0.825$ where the actor's own stated
confidence is near chance ($0.598$), and the generative readouts improve on stated confidence by $+0.07$ to
$+0.28$ across the three additional actors we audit. Against self-consistency it gains $+0.14$ to $+0.19$ on
near-deterministic actors, at $1/K$ the cost of resampling. The signal then drives two deployment modes. A real-time
gate escalates an agent's least-trustworthy calls for human review ($+0.05$ to $+0.30$ accepted-action
accuracy at $50\%$ coverage). And \emph{confidence feedback} returns the tool result together with the
score, so the agent adjusts its own next step; this lifts end-to-end task success on live-execution
benchmarks ($+0.119$ and $+0.137$, both $p{\le}10^{-4}$) and beats a random-value control where step
errors are silent ($+0.078$, $p{=}0.003$).
\end{abstract}

\section{Introduction}
Consider a customer-service or data agent running on a frontier chat API. It issues a refund tool call, a
database query, or a block of Python. If that action is silently wrong, the cost lands downstream, after
execution. The operator would like a confidence signal to gate the action---but in this, the \emph{common}
deployment setting, the two signals a deployer would reach for are both unavailable. (i)~The most direct
signal, the actor's own token log-probabilities, is hidden by the frontier chat APIs agents run on. (ii)~The
standard black-box substitute is \emph{self-consistency}: resample the actor $K$ times and measure
agreement. It collapses here, because frontier agents are \emph{near-deterministic}. The actor re-emits the
identical call on the overwhelming majority of inputs, leaving almost no sample variance to read---and it
costs $K$ frontier generations to find that out (\S\ref{sec:beats}). The actor's own verbalized
self-assessment, meanwhile, is least reliable exactly on its confident mistakes. If we cannot read the actor's
probabilities and it cannot detect its own errors, where should a confidence signal come from?

Our answer is a \emph{proxy} log-probability. A small, open-weight, log-probability-capable model runs as
a training-free \emph{sidecar} (Figure~\ref{fig:hook}): it reads the same $(\text{context},\text{schema},
\text{proposed action})$ the actor conditioned on---no gradient access, no fine-tuning, no actor internals,
no shared weights---and emits a scalar confidence. That confidence \emph{ranks} action correctness
better than either black-box baseline (\S\ref{sec:beats}). Its edge over the actor's self-assessment is
largest exactly on the actor's confident mistakes, such as hard code it cannot self-verify. And it
significantly beats self-consistency in the near-deterministic regime deployed agents occupy, at $1/K$ the
cost of resampling a frontier model. Because the auditor is a smaller open model its work is a
\emph{prefill}---one forward pass over the already-known action, no autoregressive decode---making it a
lightweight, actor-agnostic plug-in for any framework. Cost, latency, and the fail-open guarantee are in
\S\ref{sec:recipe}.

The signal is not monolithic. Which readout to trust has \emph{predictable structure}, and that structure is
what makes the sidecar both cheap and targetable. Every proxy readout is one of two types: a
\emph{generative} likelihood (teacher forcing, or request-PMI, which discounts values the scorer finds likely
\emph{regardless} of the request) or a \emph{discriminative} decision (a YES/NO verdict, tool-choice
competition). This one distinction predicts two things. (a)~\emph{Which failure mode a readout catches}: a
generative likelihood reads wrong argument \emph{values}, tracking judge-free per-parameter accuracy where
the verdict is blind to it, while the verdict reads holistic wrongness (\S\ref{sec:toolkit}).
(b)~\emph{Whether a small scorer suffices}: teacher forcing is \emph{access}-bound, equivalence-tested
against a scorer twice its size, whereas the verdict is \emph{capability}-bound and improves with scale
(\S\ref{sec:cheap}). Because the readout type localizes the error, inspecting \emph{which} readout fires does
more than rank reliability---it tells an operator what to look at, turning a scalar into a first diagnosis of
the likely failure mode. Even the safe-default fusion can be decomposed into its per-type readouts for that
diagnosis. We study function-calling and code agents: their actions are the consequential, error-prone
decisions in deployed systems, and their correctness is checkable from the call, context, and schema.

We show that such a sidecar beats the two baselines a black-box deployer would otherwise reach for
(\S\ref{sec:beats}). Read as a reliability \emph{ranking}, a small open-weight auditor's log-probability
out-predicts both the actor's verbalized self-assessment and its self-consistency, and it does so across
three actor families. The \emph{beats-self} margin is regime-graded: the verdict wins in the holistic
regime, the generative TF/PMI in argument-value regimes. We then put
the readout-type framework to work (\S\ref{sec:toolkit},\ref{sec:cheap}). Because which single readout wins
depends on the call's error locus, an orientation-corrected fusion of the family is the lowest-regret default
when that locus is unknown---as it is in deployment. A learned combiner does no better than a simple oriented
average.

Finally, we evidence two deployment modes end-to-end (\S\ref{sec:act}). The ranking drives selective
prediction on every dataset, needing only an escalation channel. And returning the score to the agent
alongside the tool result lifts end-to-end task success on two live-execution benchmarks, beating a
random-value control where step errors are silent---a distinction an unannotated control alone would have
missed. Throughout, claims are regime-graded and scoped to that class of actions.

\section{Background and related work}\label{sec:related}
Confidence estimation for LLMs differs mainly in the \emph{source} of the signal. Every existing
source is unavailable or unreliable once the actor is a deployed, black-box, near-deterministic agent.
\textbf{Actor-internal probabilities.} P(True)/self-evaluation reads the model's probability on its
\emph{own} answer~\citep{kadavath2022}---intra-model, and inaccessible behind a frontier chat API.
\textbf{Actor verbalized confidence} asks the model to \emph{state} a number~\citep{lin2022teaching,
tian2023just,xiong2024can}; it is reliable only where the model is well-calibrated, and least reliable
exactly on its confident errors. \textbf{Actor sampling/agreement}---self-consistency~\citep{wang2023selfconsistency},
SelfCheckGPT~\citep{manakul2023selfcheckgpt}, semantic entropy~\citep{kuhn2023semantic,farquhar2024detecting}%
---needs multiple actor samples and collapses under the near-determinism of deployed frontier agents
(\S\ref{sec:beats}). All three read the \emph{actor}; none survives the black-box, near-deterministic regime.

\begin{sloppypar}
\textbf{A second model as the signal source.} LLM-as-judge~\citep{zheng2023judging} and
cross-examination~\citep{cohen2023lm} bring in another model but read its \emph{words}---a verbalized
verdict; weak-to-strong generalization~\citep{burns2024weak} and learned process/outcome
verifiers~\citep{lightman2024verify} require a \emph{training} step; guardrail
classifiers~\citep{inan2023llamaguard} screen content, not action correctness.
\textbf{Cross-model likelihood as a metric.} Scoring text by a \emph{second} model's token
log-probabilities is established in generation metrics: BARTScore casts evaluation \emph{as} generation and
reads a scorer model's likelihood~\citep{yuan2021bartscore}, GPTScore steers the same quantity by a
natural-language instruction~\citep{fu2024gptscore}, while learned MT metrics instead regress on human
judgments~\citep{rei2020comet}; probability-curvature tests read likelihood \emph{geometry} to detect
machine-generated text~\citep{mitchell2023detectgpt,bao2024fastdetectgpt}. All of these score
\emph{prose}---its fluency, factuality, or provenance---never whether a \emph{tool call will execute
correctly}. And none is organized by which failure mode the readout catches. Our request-PMI readout follows the
domain-conditional-PMI correction introduced against surface-form competition in answer
scoring~\citep{holtzman2021surface}, reweighting a likelihood by its context-free prior. \textbf{Calibration and oversight.} Post-hoc calibration sharpens
a classifier's \emph{own} probabilities~\citep{guo2017calibration}, presuming the access we lack; monitoring an
untrusted model with a weaker trusted one is the control-protocol setting~\citep{greenblatt2024aicontrol}, to
which we contribute a concrete training-free monitor signal.
We build on standard deployment machinery rather than claim it: accept-if-passes execution-feedback
repair~\citep{chen2024selfdebug,chen2023codet,ni2023lever} (self-correction has known limits without an
external check~\citep{huang2024large}), selective prediction /
risk--coverage~\citep{elyaniv2010foundations,geifman2017selective}, and cost-driven
cascades~\citep{chen2023frugalgpt}, which our two deployment modes instantiate.
\textbf{The gap we fill:} inference-time, training-free, black-box-compatible \emph{action} verification
from a second model's log-probabilities---not its words, not the actor's own---organized by a failure-mode
map over generative/discriminative readout types.
\end{sloppypar}

\section{The proxy-confidence sidecar}\label{sec:setup}

\begin{figure}[tb]\centering
\includegraphics[width=0.86\linewidth]{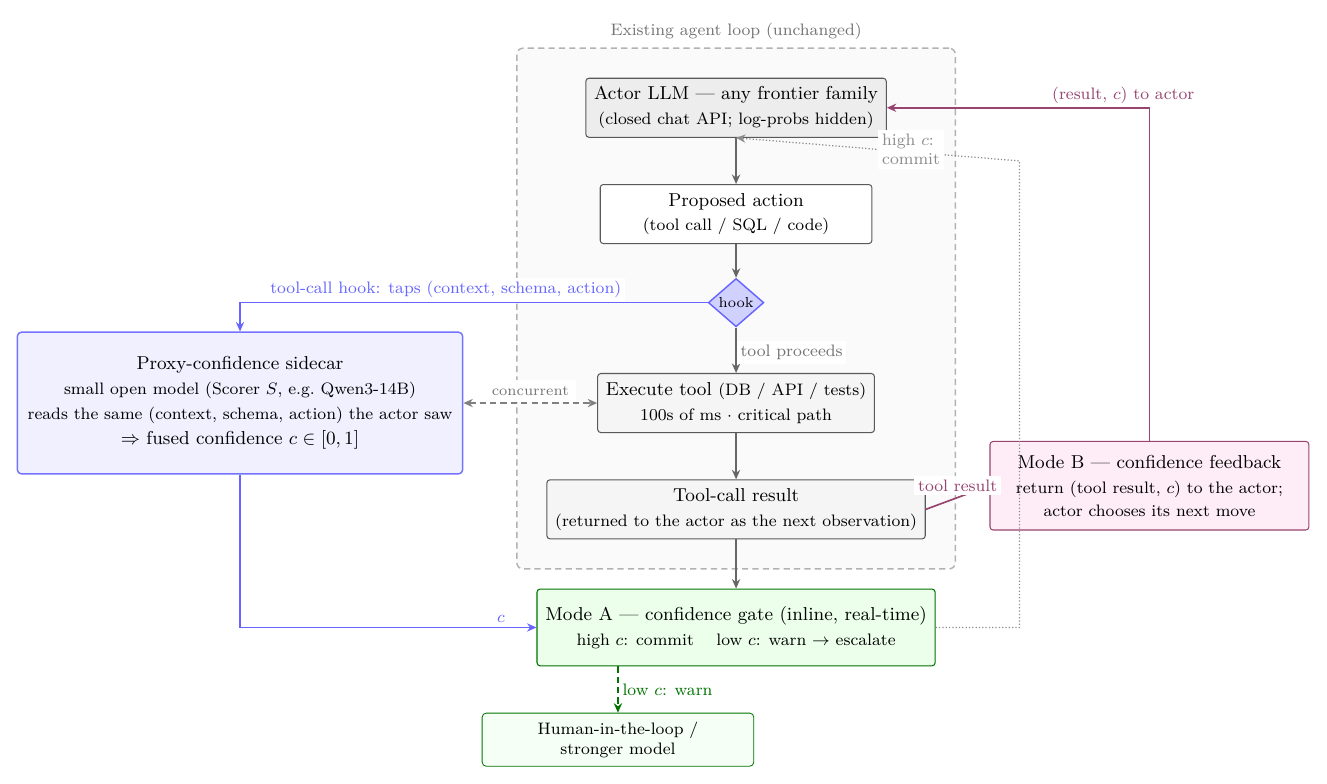}
\caption{\textbf{The proxy-confidence sidecar: a training-free plug-in hook.} The sidecar (dashed) taps the
agent's tool-call hook, reads only $(\text{context},\text{schema},\text{action})$---no logits, weights, or
fine-tuning---and emits a confidence $c$, scored in parallel with tool execution. Drawn here are the two
in-loop modes: a real-time oversight gate (Mode~A: high $c$ commit, low $c$ escalate to a human) and
\emph{confidence feedback} (Mode~B: return the tool result together with $c$ to the actor, which chooses its
next move). Details in \S\ref{sec:act}--\S\ref{sec:recipe}.}
\label{fig:hook}
\end{figure}

A black-box actor $\mathcal{A}$ emits an action $o$ and exposes no log-probabilities. A log-probability-capable auditor $\mathcal{S}$ (equivalently, the
\emph{scorer}) reads only $(\text{context},\text{tool schema},o)$---\emph{the same information the actor
conditioned on}---and computes a scalar confidence (Table~\ref{tab:signals}) at temperature~$0$. Our primary
auditor is a \emph{small, open-weight} model (Qwen3-32B or 14B, or Gemma-3-27B); this is the cheap-sidecar
setting the cost argument (\S\ref{sec:recipe}) rests on. For the one hard-code case we instead score with a
large open-weight model of different lineage than the actor, to rule out same-family self-preference bias;
we write $\mathcal{S}_{\text{L}}$ for that scorer and flag it explicitly at that result (\S\ref{sec:beats}).
\textbf{Actors are de-identified.}\footnote{This paper studies a \emph{property of the black-box deployment
setting}---that an actor's stated confidence is uninformative about its own confident errors---not the
standing of any particular vendor's model. Per-model reliability figures are therefore reported under
neutral, de-identified labels, and we make no comparative claim about any provider. All actors are recent
instruction-tuned models scored by the same readouts under identical prompts and labels. No claim here needs
an actor's identity to check: each is either within-actor or holds across \emph{all} actors.}
We study five actors across three families: \textbf{A1} and \textbf{A2}, two sizes of a widely deployed
closed frontier family; \textbf{B}, a frontier open-weight model; and \textbf{C-120b}/\textbf{C-20b}, two
sizes of a second open-weight family. Most analyses use A1 and A2 and add the others to test cross-family
generalization (\S\ref{sec:beats}); $\mathcal{S}_{\text{L}}$ is of the same lineage as actor~B, so for that
one result the scorer and an actor share a family, which is why it is never used to score B. The baseline a black-box
deployer would otherwise use is $\mathcal{A}$'s own verbalized self-assessment (and, in \S\ref{sec:beats},
its self-consistency). \textbf{Dataset selection.} We fix a single a-priori inclusion rule: a benchmark
qualifies iff action correctness is checkable by execution/AST/gold-match \emph{without an LLM judge}. We
then take the most widely-used qualifying benchmark in each action category, spanning three model families:
function-calling (BFCL~\citep{bfcl2024,patil2024gorilla}, xLAM/APIGen~\citep{liu2024apigen},
NESTFUL~\citep{basu2024nestful}), interactive coding (AppWorld~\citep{trivedi2024appworld}, InterCode
\textsc{nl2bash}~\citep{yang2023intercode}), code generation
(LiveCodeBench~\citep{jain2025livecodebench}, APPS~\citep{hendrycks2021apps},
CodeContests~\citep{li2022alphacode}), and text-to-SQL (BIRD~\citep{li2023bird}). We report on \emph{every}
qualifying benchmark we ran rather than omitting any, including the one regime where the method
\emph{fails}---tau2-bench~\citep{yao2024taubench}, which is omission-dominated and retained as a scope note
(\S\ref{sec:limits}). Labels are \emph{execution/AST/gold-match based} per dataset
(AppWorld unit tests; BFCL state/response checkers and AST-match; NESTFUL/xLAM gold-call matching; public
tests for code; shell-state and file-match for InterCode)---never LLM-judged.

\paragraph{Two primitives.} Every readout is one of two types, and this distinction organizes the paper. A
\emph{generative} readout scores the \emph{likelihood} the auditor assigns to the action, either raw (teacher
forcing, TF) or request-contrasted (request pointwise mutual information, request-PMI). The contrast matters
because raw likelihood rewards generic argument values: PMI subtracts the same value's log-probability with
the \emph{request withheld}, keeping only the evidence the request supplies. A
\emph{discriminative} readout is instead the auditor's
\emph{posterior over a discrete label set}---over correctness (the verdict) or over function identity
(tool-choice competition). The type predicts both the failure mode a signal catches (\S\ref{sec:toolkit})
and whether a small scorer suffices (\S\ref{sec:cheap}); Table~\ref{tab:signals} lists the four instances.
The verdict needs only generated-token log-probabilities (any chat API that exposes them), whereas TF/PMI
require prefill/echo access. We elicit the verdict two ways and report which throughout.
\emph{Generated-token} elicitation reads the log-probability of the scorer's own next token after the ``is
this correct?'' prompt. \emph{Matched} elicitation instead forces YES and NO and renormalizes over the pair;
because that is the same mechanism as teacher forcing, it is confound-free when we compare the verdict
against the generative signals. The two mostly agree, diverging where free-form generation carries
formatting cues the forced pair strips out. The divergence is sharpest on xLAM: there the generated-token
verdict ties the actor while the matched verdict is near-chance, so the xLAM claim rests on TF/PMI rather
than the verdict.

\paragraph{Metric and reporting.} The metric is AUROC for predicting action correctness; significance
statements use paired bootstrap over items. We report two quantities and mark them by case: uppercase
``$P$'' is a paired-bootstrap exceedance probability (the fraction of bootstrap resamples in which the proxy
signal out-predicts the baseline), and lowercase ``$p$'' is a frequentist test $p$-value.\footnote{The
frequentist test is matched to the contrast: paired-McNemar for a change in end-to-end accuracy over the
same items (Mode~B, Table~\ref{tab:inline}); Holm-corrected across arms for the multi-arm self-review
family (Table~\ref{tab:active}); and a paired-bootstrap TOST for equivalence claims (\S\ref{sec:cheap}).}
Signals are
\emph{definable} on different action types: value-token TF/PMI need a single forced call with JSON arguments,
while the verdict applies to any action. Each table therefore reports the
datasets/actors/scorers for which its signal is defined and relevant (full coverage matrix,
App.~\ref{app:coverage}, Table~\ref{tab:coverage}). Cross-dataset heterogeneity is high ($I^2{=}90\%$), so we report
\emph{per-dataset} AUROC and never pool \emph{absolute} AUROC. The only pooled numbers are \emph{paired}
contrasts---a within-item sign test of whether the proxy signal out-predicts the same-item baseline---for
which level differences cancel. tau2-bench is used only as a scope note, not a per-call benchmark
(\S\ref{sec:limits}).

\begin{table}[t]
\centering\small
\caption{\textbf{The four proxy readouts and the failure mode each targets.}}
\label{tab:signals}
\setlength{\extrarowheight}{5pt}
\begin{tabular}{p{1.95cm}>{\raggedright\arraybackslash}p{3.45cm}>{\raggedright\arraybackslash}p{1.9cm}}
\toprule
Signal & Readout & Targets \\
\midrule
\textbf{Verdict} & $P(\text{YES})$ on ``is this correct?'', renormalized over YES/NO & coarse
tool/code correctness \\[5pt]
\textbf{Teacher forcing} & per-token $\log P$ of $\mathcal{A}$'s exact call, localized to argument
\emph{value} tokens & wrong argument \emph{values} \\[5pt]
\textbf{Request-PMI} & \mbox{$\log P(o\mid\text{full})$} \mbox{${-}\log P(o\mid\text{req.\ withheld})$}, on value tokens &
groundedness of args in request \\[5pt]
\textbf{Tool-choice competition} & does $\mathcal{S}$ prefer $\mathcal{A}$'s function over $k$ siblings? &
wrong \emph{function} choice \\
\bottomrule
\end{tabular}
\end{table}

\section{The sidecar beats the deployer's two baselines}\label{sec:beats}
The deployer's two black-box options both fail in the deployment regime, and the proxy signal beats each.
\textbf{(1) vs.\ the actor's self-assessment.} Every proxy readout (\S\ref{sec:setup}) reads the scorer's
log-probabilities rather than its words. A within-model \emph{mechanism ladder} (Figure~\ref{fig:ladder})
confirms this is where the signal comes from, and quantifies the payoff. On the \emph{same} Qwen3-32B we change only the readout---from a
verbalized $0$--$100$ confidence to the verdict token's log-probability, identical model and prompt
content---and AUROC rises $+0.09$ to $+0.18$ on $5/6$ datasets (exceedance $>0.99$; miss\_param, verdict-blind
by construction, is the exception at $+0.02$). Capability is held exactly constant, so the gain is
attributable to \emph{access}, not scale. The intuition is that a verbalized number is a lossy projection of a
model's internal uncertainty onto a few output tokens, and it is lossy at any scale. Direct
log-probability access therefore recovers signal that added capability in the actor cannot---which is why an
auditor smaller than the actor can still out-predict its self-report. The payoff follows. Pooled over the five clean datasets where actor self-assessment
was logged (AppWorld, BFCL base/long, miss\_func, xLAM; $n{=}2439$; the verdict-blind miss\_param is
excluded), the smaller instrumented scorer's verdict out-predicts the frontier actor's own self-assessment
(pooled \emph{paired} contrast $+0.022$ AUROC, CI$_{95}$ $[{+}0.002,{+}0.043]$, $P{=}0.98$---a within-item
sign test of direction, not a pooled absolute AUROC). That edge is \emph{small but significant} overall, and
much larger on the individual regimes where the actor is most overconfident. The starkest such regime is
hard code, which we show as a
separate cross-model \emph{judge-solve asymmetry} (here the domain-agnostic neutral-verdict scorer
$\mathcal{S}_{\text{L}}$, not the sub-frontier Qwen, scoring actor~A1's LiveCodeBench solutions, $n{=}139$):
the proxy verdict reaches $\mathbf{0.825}$ while the actor's own stated confidence is near chance at $0.598$
(paired $\Delta{+}0.213$, CI$_{95}$ $[{+}0.13,{+}0.33]$, $P{=}1.0$). Judging code is easier than writing it
and a stated number does not recover that asymmetry: a competent actor can still be an uninformative reporter
on its own hardest outputs---the silent-failure regime a guardrail is for, and a property of
\emph{self-report as a signal source} rather than of any one model.
That ranking \emph{transfers}: a calibrator fit on other datasets and evaluated leave-one-dataset-out on
held-out LiveCodeBench still scores $0.836$ and beats the actor by $+0.222$, so the gap is not an
overfit artifact (App.~\ref{app:sc} reconciles the three LiveCodeBench verdict figures).

\textbf{(2) vs.\ self-consistency---and why the deployment regime is where proxy scoring wins.} The other
black-box baseline is \emph{actor} self-consistency (SC): sample the actor $K{=}5$ times and use output
agreement, at $K{\times}$ the actor's cost. SC's AUROC is governed by the actor's \emph{own} sample entropy, so
proxy-vs-SC is partly a comparison of actors---which is precisely why proxy scoring wins where agents are
deployed. On the near-deterministic actors that dominate deployment (family~A, $94$--$96\%$ of
resampled calls identical) SC has almost no variance to read and falls to $0.57$--$0.62$, while the
proxy generative signal holds $0.72$--$0.77$ (paired PMI$-$SC $=+0.19$ A1, $+0.14$ A2, both
CI$_{95}$ excluding $0$; SC's AUROC decays with sample unanimity at $r{=}{-}0.96$ across five actors,
App.~\ref{app:sc}). Near-determinism is a precondition SC needs \emph{absent}, not an actor defect.
We do \emph{not} claim a blanket win. On higher-entropy actors (family~C) SC
is competitive ($0.76$--$0.81$), and there neither the best single readout nor a fusion of the applicable
readouts (oriented fusion, \S\ref{sec:toolkit}) cleanly beats it. What is unconfounded and \emph{universal} is the cost asymmetry---one auditor pass vs.\ $K$ frontier
generations---plus the significant AUROC win in that regime. (On code, cluster-agreement of the samples \emph{on the public tests} scores $0.72$--$0.84$, but
that runs the test oracle the auditor never sees. Made execution-free to match the auditor's no-test access
it drops to $0.52$--$0.64$, and the proxy verdict beats it---LiveCodeBench $0.851$ vs.\ $0.623$;
App.~\ref{app:sc}.)

\textbf{Auditing generalizes across families; the beats-self margin is regime-graded.} We add the three
actors outside family~A on BFCL (B; C-120b/C-20b), scored by the same Qwen/Gemma auditors---a fully
cross-family pairing with no shared weights. The proxy verdict predicts every actor's turn correctness
at AUROC $0.64$--$0.77$, and is cross-scorer-stable (Qwen${\approx}$Gemma within ${\approx}0.02$; Table~\ref{tab:grid}).
So \emph{auditing} generalizes. The \emph{beats-self} edge is regime-graded, and the two readout types split
it cleanly along the failure-mode map we develop next (\S\ref{sec:toolkit}). The discriminative
verdict beats self on family~A's holistic BFCL regime, but only \emph{ties} on B and C, which
self-assess well there ($0.66$--$0.77$). The generative TF/PMI beats self on the
argument-value regime for \emph{all three} of the remaining actors: teacher-forcing
each of their xLAM calls through the proxy Qwen3-32B, request-PMI predicts correctness at
$0.72$--$0.79$ and improves on the actor's own stated confidence by $+0.07$ to $+0.28$, all significant
(Table~\ref{tab:crossfam}). The margin is large where the actor states confidence near chance
($0.51$--$0.53$), and smaller but still significant where the actor self-assesses well at $0.65$
yet still loses to PMI. Whether the proxy signal beats an actor's \emph{verbalized} self-assessment
thus depends on the actor's self-calibration and on the regime. We report that as a family/regime-graded
effect rather than universal dominance. Because the auditor is open, it also admits an epistemic readout a
black-box pipeline cannot: ensemble disagreement between two different-lineage scorers ($0.70$--$0.80$,
near-free; App.~\ref{app:signals}).

\emph{Taken together:} auditing generalizes. A proxy log-probability ranks correctness across three
unrelated actor families with no shared weights, and across action types (tool calls and code;
\S\ref{sec:beats},~\S\ref{sec:act}). Beating the actor's \emph{own} confidence, by contrast, is
regime-graded---strongest exactly where the actor is blind to its own errors. The sidecar is not a universal
oracle. It is a guardrail whose edge grows with the actor's error rate, scoped to actions whose correctness
is decidable by reading them (\S\ref{sec:limits})---precisely the regime a deployer cares about.

\begin{table}[t]
\centering\small
\caption{\textbf{PMI improves on stated confidence for all three actors outside family~A.} Proxy signals
scoring each actor's xLAM calls through Qwen3-32B; best proxy in bold. Actors are de-identified
(\S\ref{sec:setup}); the comparison of interest is proxy-vs-self \emph{within} each row, not between rows.}
\label{tab:crossfam}
\setlength{\tabcolsep}{4pt}
\begin{tabular}{@{}lccccc@{}}
\toprule
Actor (family; $n$) & TF & \textbf{PMI} & verd. & self & PMI$-$self \\
\midrule
C-120b (family C; 775) & 0.735 & \textbf{0.794} & 0.526 & 0.515 & $+0.279$ \\
C-20b (family C; 771)  & 0.743 & \textbf{0.793} & 0.552 & 0.534 & $+0.255$ \\
B (family B; 1261)     & 0.636 & \textbf{0.721} & 0.546 & 0.648 & $+0.071$ \\
\bottomrule
\end{tabular}
\\[2pt]{\footnotesize PMI$-$self CI$_{95}$: C-120b $[{+}.222,{+}.338]$, C-20b
$[{+}.197,{+}.314]$, B $[{+}.021,{+}.121]$.}
\end{table}

\section{The signal family and its failure-mode map}\label{sec:toolkit}
Log-probability access buys more than one signal. We state a failure-mode map \emph{a priori}---which
readout should win on which action type---and then test it (Table~\ref{tab:hero}; the same matrix as a
heatmap is in App.~\ref{app:signals}, Fig.~\ref{fig:heatmap}). The champion signal marches predictably from the
\emph{verdict} on coarse holistic regimes to \emph{teacher forcing} on argument-heavy ones to
\emph{tool-choice competition} on wrong-function risk.

\begin{table}[t]
\centering\small
\caption{\textbf{The champion signal tracks the error locus.} Per-signal AUROC (32B scorer) vs.\ the
actor's self-assessment, best single signal in bold.}
\label{tab:hero}
\setlength{\tabcolsep}{3.5pt}
\begin{tabular}{lccccccc}
\toprule
Dataset & locus & verd. & TF & PMI & actor & fuse & $P^\star$ \\
\midrule
AppWorld & holistic & \textbf{0.894} & n/a & n/a & 0.788 & n/a & .996 \\
BFCL base & holistic & \textbf{0.733} & 0.633 & n/a & 0.703 & 0.720 & .945 \\
BFCL long & holistic & \textbf{0.719} & 0.640 & n/a & 0.682 & 0.700 & .968 \\
BFCL v3 & args & 0.711 & 0.549 & \textbf{0.714} & 0.636 & 0.700 & .950 \\
parallel & args & 0.661 & 0.653 & 0.662 & 0.577 & \textbf{0.695} & .980 \\
NESTFUL & args & 0.594 & \textbf{0.643} & 0.554 & 0.620 & \textbf{0.655} & $.840$ \\
miss\_param & args & 0.517 & \textbf{0.573} & n/a & 0.528 & \textbf{0.697}$^{\P}$ & .950 \\
miss\_func & function & 0.592 & n/a & n/a & 0.624 & 0.681$^{\parallel}$ & --- \\
\midrule
\emph{xLAM}$^{\natural}$ & easy & \emph{0.720} & \emph{0.750} & \emph{0.760} & \emph{0.710} & n/a & --- \\
\bottomrule
\end{tabular}
\\[2pt]
{\footnotesize \emph{fuse}: oriented rank-average of verdict$+$TF; $P^\star$: paired-bootstrap exceedance
(best single vs.\ actor); n/a: signal undefined for that action type, not a skipped run.
$^{\parallel}$tool-choice competition (not verdict$+$TF). $^{\P}$learned fusion, argument-bearing subset
($n{=}565$). $^{\natural}$xLAM verdict ties the actor only under generated-token elicitation; under matched
elicitation it is near-chance, so the xLAM claim rests on TF/PMI (App.~\ref{app:cap14}).}
\end{table}

\paragraph{Verdict---holistic correctness (the default).} Where correctness is adjudicable on-face, the
one-token verdict is cheapest and best (AppWorld $0.894$ vs.\ actor $0.788$, $P{=}.996$; BFCL base/long),
and on AppWorld beats a trajectory-length baseline the actor's self-report does not---diagnostic, not a
length proxy.

\paragraph{Teacher forcing / request-PMI---when argument \emph{values} carry the risk.} Here the holistic
verdict is too coarse. On NESTFUL, TF ($0.643$) beats the verdict ($0.594$, $P{=}.98$). On
miss\_param (a required argument withheld) the verdict is at chance ($0.517$) while TF recovers
$0.573$ and beats the actor, and the ranking is actor-invariant (A2: TF $0.604 >$ verdict $0.552 >$ self
$0.594$). A \emph{judge-free} anchor confirms the mechanism: against deterministic per-parameter accuracy, TF's
correlation rises to $\rho{=}0.41$ as the target sharpens to argument values, while the verdict collapses to
$\rho{=}0.02$---for both actors (App.~\ref{app:signals}, Fig.~\ref{fig:perparam}). On short structured calls
(BFCL v3) request-PMI ($0.714$) ties the verdict. And for wrong-\emph{function} risk (miss\_func) a
counterfactual over $k{=}4$ sibling tools reaches $0.681$ where the verdict is weak ($0.592$).

\paragraph{Fusion---additive when signals are complementary, and orientation is mandatory.}\begin{sloppypar}We
write \emph{oriented fusion} for the orientation-corrected rank-average of whichever readouts are defined on
a given call---in practice verdict$+$TF, plus PMI or tool-choice where they apply. (A larger
\emph{full-battery fusion} of all 16 single-pass readouts is an appendix probe, App.~\ref{app:battery}.) Verdict and TF
are nearly orthogonal ($\rho{\le}0.36$), so rank-averaging helps most where they catch \emph{different}
errors: miss\_param ($0.517/0.573{\to}0.697$ learned, ${\approx}0.62$ plain rank-average, beating either and
the actor), parallel ($0.695$), NESTFUL ($0.655$). It is not free: where signals share errors it does not
exceed the best single readout, and PMI on top of verdict$+$TF \emph{hurts} on nested args
($0.655{\to}0.625$). Two rules govern it. First, the combiner barely matters---a learned logistic is
indistinguishable from an orientation-corrected rank-average, so we claim no trained combiner
(App.~\ref{app:battery}). Second, orientation is essential, and free: some TF signals are
\emph{anti}-correlated with correctness in a regime. This happens when the fluent, high-likelihood
completion is the wrong one---the scorer finds a generic or default-valued call likelier than a correct call
needing an unusual argument, so higher TF flags the easy-but-wrong action. A na\"ive average that fails to
flip such a signal can fall below chance, so we fit each sign on a held-out fold.\end{sloppypar}

Which signal to trust is therefore not a mystery. The failure-mode map makes it predictable from where the
error can live---generative for argument values, discriminative for holistic wrongness. Two
operational questions follow, and the same generative/discriminative split answers both: does the sidecar
need a large scorer, and can it run cheaply?

\section{Which readout, how cheap, and when to specialize}\label{sec:cheap}
The generative/discriminative type answers the first of the two questions \S\ref{sec:toolkit} closed on, and
the answer makes the sidecar cheap.

\begin{sloppypar}
\textbf{Teacher forcing $=$ access-bound: a 14B suffices (equivalence-tested).} We re-score every dataset with
Qwen3-14B (bf16)---half the 32B's parameters, every signal through the \emph{identical} prefill on
both scorers. The teacher-forcing readout is \emph{capability-invariant}: on all six matched cells the 14B
tracks the 32B within $|\Delta|{\le}0.015$ AUROC (on xLAM the 14B is fractionally ahead), and a
paired-bootstrap TOST equivalence test ($\delta{=}0.05$) passes on all $6$ (App.~\ref{app:cap14},
Fig.~\ref{fig:split}). Reading an action's raw likelihood needs log-probability \emph{access}, not scale: the
quantity is intrinsic to the forward pass over low-entropy value tokens. Request-PMI contrasts two
prefills and is more scale-sensitive than raw TF ($-0.04$--$0.05$ at 14B on the two cells we test), so we treat
the 14B-suffices claim as specific to teacher~forcing. \textbf{Discriminative $=$
capability-sensitive:} the verdict falls on the other side. Elicited the same way on both scorers
(confound-free), the 32B verdict significantly beats the 14B on $6$ of $9$ matched dataset$\times$actor cells
(90\% CI excludes 0; none favor the 14B). The three exceptions are the two NESTFUL cells and AppWorld, where
trajectory correctness is legible even to the small scorer; Table~\ref{tab:cap14} shows representative cells. The mechanism is a difference in kind: a generative likelihood is intrinsic to the forward pass, whereas the
verdict must instantiate a \emph{learned decision boundary} over an abstract correctness latent---a
meta-competence that scales with capability. With $n{=}2$ sizes we report this as a capability \emph{trend},
not an invariance law. \textbf{Deployment consequence:} provision a $14$B scorer for access-bound
teacher forcing, and add the $32$B when the verdict, request-PMI, or a fusion including them carries
weight.
\end{sloppypar}

\textbf{Robustness: not one pairing.} A $2{\times}2{\times}3$ grid (A1/A2 $\times$
Qwen3-32B/Gemma-3-27B $\times$ AppWorld/BFCL~base/long; Table~\ref{tab:grid}) shows the verdict is
not a Qwen artifact. The two different-lineage scorers agree within ${\approx}0.02$ AUROC on BFCL (and
${\approx}0.06$ on AppWorld), and both beat both actors' self-assessment (exceedance $>0.99$ against the
stronger A2). The lone exception is AppWorld$\times$A2, a strong self-assessor on its own long code
trajectory. The cross-family evidence for families~B and~C (Tables~\ref{tab:crossfam},~\ref{tab:grid}) was presented
in \S\ref{sec:beats}.

\section{From confidence to action: two deployment modes}\label{sec:act}
The ranking is directly useful in two modes (Figure~\ref{fig:hook}). \textbf{Mode A---selective
prediction---is the uniform, always-available use.} Ranking calls by proxy confidence and
\emph{accepting only the most-confident fraction} (abstaining/escalating the rest) improves accepted-action
accuracy on \emph{every} dataset---a direct consequence of AUROC${>}0.5$ (per-dataset numbers in
Table~\ref{tab:selective}, App.~\ref{app:selective}). At
$50\%$ coverage on \emph{natural} errors, accepted accuracy rises $+0.05$ (BFCL), $+0.16$ (xLAM), and $+0.095$
on tau2 (scored at the trajectory level against end-to-end task success, not per call). On
the injected-error \emph{needle} construction---which corrupts exactly one call per trajectory and asks the
signal to localize it---accuracy rises $+0.22$--$0.30$. The risk--coverage area is far below random
abstention everywhere. Needing only an abstain/escalate channel and no self-repairing actor, Mode~A is the
one use that works on any black-box agent. It can also buy efficiency rather than accuracy: at a fixed audit
budget of one call per trajectory, ranking by proxy confidence repairs the injected error $0.92$ of the time
on xLAM and $0.65$ on NESTFUL, versus $0.42$/$0.40$ for a random-call budget ($p{<}10^{-4}$, $p{=}0.008$), so
it decides \emph{where} to spend review even where blanket review saturates. (Handing the actor its own
confidence \emph{post hoc} to self-correct helps only where it cannot otherwise self-verify the call;
App.~\ref{app:selfreview}. tau2 probes an orthogonal \emph{omission}-dominated regime and is a scope note,
\S\ref{sec:limits}.)

\textbf{Mode B---confidence feedback: return $(\text{result},c)$ and let the actor act.} The most direct use
of the hook is the one Figure~\ref{fig:hook} draws: append the confidence to the tool result the actor reads
next, and let it choose its own next move---no gate, no human. With an A2 actor, oriented fusion raises
end-to-end task success on both live-execution benchmarks:
\textsc{nl2bash}~\citep{yang2023intercode} $0.399{\rightarrow}0.518$ ($+0.119$; $29$ tasks rescued vs.\ $6$
broken, exact $p{=}10^{-4}$; $n{=}193$) and AppWorld \texttt{test\_challenge} $0.595{\rightarrow}0.731$
($+0.137$; $100/43$, $p{<}10^{-4}$; $n{=}417$).

Two controls then \emph{separate} the benchmarks, and that split is the point. Announcing an audit while
withholding the number does nothing (\textsc{nl2bash} $-0.010$, $p{=}0.87$); an independent $U(0,1)$ draw is
likewise flat ($+0.041$, $p{=}0.28$), and fusion beats that draw by $+0.078$ ($19/4$, $p{=}0.003$)---so here
the annotation's \emph{content} matters, not merely its presence. AppWorld inverts this: the random stream
matches fusion ($+0.005$, $46/44$, $p{=}0.92$) and the displayed signal sits at chance (AUROC
$0.39$--$0.47$ vs.\ $0.64$). The reason is diagnostic---AppWorld's Python failures mostly \emph{raise}, so the
traceback already carries the diagnosis, whereas on \textsc{nl2bash} half of all wrong steps emit no error
text at all. Mode~B
therefore earns its keep only where step errors are silent---and a random-value arm is what tells an operator
which of the two regimes they are in (App.~\ref{app:selfreview}).

\section{Deploying the sidecar: cost, latency, and the safe default}\label{sec:recipe}
\paragraph{Why the sidecar is cheap: the scorer's work is a prefill.} The confidence readouts are dominated
by \emph{prefill}, not autoregressive decode. Teacher-forcing the actor's \emph{already-known} action needs
one forward pass over known tokens with \emph{zero} generation; request-PMI needs two; the verdict needs one
prefill plus a single token. The scorer never pays the long decode the actor already paid, so a $14$B pass
over the shared context is cheaper than the actor's own generation---and than the $K$ frontier generations SC
needs. ``Lightweight'' here means per-call compute; the sidecar is still a second model to host. A
generative-only deployment is cheapest, since a $14$B suffices for the access-bound signals, while a
verdict/fusion gate provisions the $32$B (\S\ref{sec:cheap}).
\emph{Measured} on the deployed $14$B fp8 sidecar at realistic full-context length (one L40S GPU,
Table~\ref{tab:latency}), a scored call takes $123$--$159$\,ms at the median ($p99{\le}179$\,ms). That is a
single-digit-percent addition to the
$100$s-of-ms a real tool call (DB query, API round-trip) already costs, and it is hidden entirely when the
scorer runs concurrently with execution. Scoring is triggered by the framework's pre-execution tool-call hook,
which already holds the $(\text{context},\text{schema},\text{action})$ tuple, so it needs \emph{no extra actor
calls}, adds ${\approx}0$ to the Mode~A critical path, and fails open on timeout so the guardrail can never
stall the agent. Because the hook fires per call, scoring can be restricted to high-stakes calls: as a design
policy (not a measured cost-saving) a static allowlist of writes/refunds/deletes plus a free-form-argument
heuristic routes argument-bearing calls to TF/PMI. The \emph{evidenced} selectivity is that the signal tells
you \emph{where to spend} a fixed review budget once calls are scored (\S\ref{sec:act}).

\paragraph{Oriented fusion is the regime-agnostic safe default.} The failure-mode map
(\S\ref{sec:toolkit}) says \emph{which} single signal wins in each regime. But exploiting it requires
knowing the error locus \emph{a priori}, which is exactly the deployment knowledge a black-box operator
usually lacks. Worse, every single signal has a \emph{catastrophic regime}. The verdict is near-blind
where a required argument is withheld (miss\_param $0.517$, at chance), and teacher forcing is weak on the
coarse holistic calls the verdict handles best (BFCL v3 $0.549$ vs.\ the $0.714$ optimum there). Committing
to either one blind therefore risks a large loss. We quantify this as \emph{worst-case regret}: the AUROC
gap between a fixed policy and the best-achievable signal in each regime. Reading the oriented fusion caps
worst-case regret at $\mathbf{0.019}$ AUROC (mean
$0.007$), versus $\approx0.17$--$0.18$ for either single-signal policy (breakdown in
Table~\ref{tab:regret}, App.~\ref{app:regret}).

\paragraph{Fusion or a single signal? A practitioner's rule.} The choice is \emph{risk management}, not
performance maximization: default to oriented fusion whenever you cannot predict, per call, where the error
will be---the common black-box case. Fusion is not the highest-\emph{ceiling} readout in any regime but the
lowest-\emph{floor} one, the only choice independent of the call pattern. Drop to a single signal only when you know the error locus a
priori: a \emph{verdict} for holistic correctness, \emph{teacher-forcing}/\emph{request-PMI} when argument
\emph{values} carry the risk, \emph{tool-choice competition} for wrong-\emph{function} routing. Each is
cheaper, and ties or edges fusion where you are right about the locus. The deployable edge
tracks the actor's error rate. It is largest on error-prone tool/code regimes (AppWorld $+0.11$, BFCL,
NESTFUL, miss\_param), and slightly negative only for the verdict on easy xLAM ($-0.06$, where TF/PMI still
lead)---exactly the calls the actor already gets right, which least need a guardrail.

\section{Limitations and scope}\label{sec:limits}
Our claims are scoped to actions whose correctness is decidable by \emph{reading} them---for a tool call,
whether it matches the request and schema. Where the deciding evidence is an external fact absent from the
context, we claim nothing. One hazard carries into deployment: a readout's \emph{sign} is regime-dependent.
Teacher forcing anti-correlates with correctness wherever the fluent completion is the wrong one
(\S\ref{sec:toolkit}), so orientation must be fit on held-out labels per regime and never ported; an
unvalidated sign gives an anti-correlated gate, not merely a weak one. The \emph{beats-self} edge is
regime/family-graded, not universal---on BFCL outside family~A
the verdict only ties (\S\ref{sec:beats})---and a second strong \emph{closed} frontier family is
untested. The cost claim rests on a $1/K$-generations plus smaller-model argument and single-GPU latency
(Table~\ref{tab:latency}); multi-tenant serving is unprofiled and scorer size spans only a $2\times$ ratio
($\le$8B untested). Learned fusion does
\emph{not} beat an orientation-corrected rank-average (App.~\ref{app:battery}), so we claim no trained
calibrator. Finally, Mode~B is regime-dependent: it beats a random-value control only where step errors are
silent (\S\ref{sec:act}), so it warrants a scorer-validity check Mode~A does not.
\bibliographystyle{plainnat}
\bibliography{main}

\clearpage
\appendix
\section{Mechanism ladder: access, not scale}\label{app:ladder}
The within-model comparison behind the ``access, not scale'' claim of \S\ref{sec:beats}: on a fixed scorer,
only the readout changes (verbalized confidence $\to$ verdict-token log-probability), isolating access from
capability.

\begin{figure}[h]\centering
\includegraphics[width=0.82\linewidth]{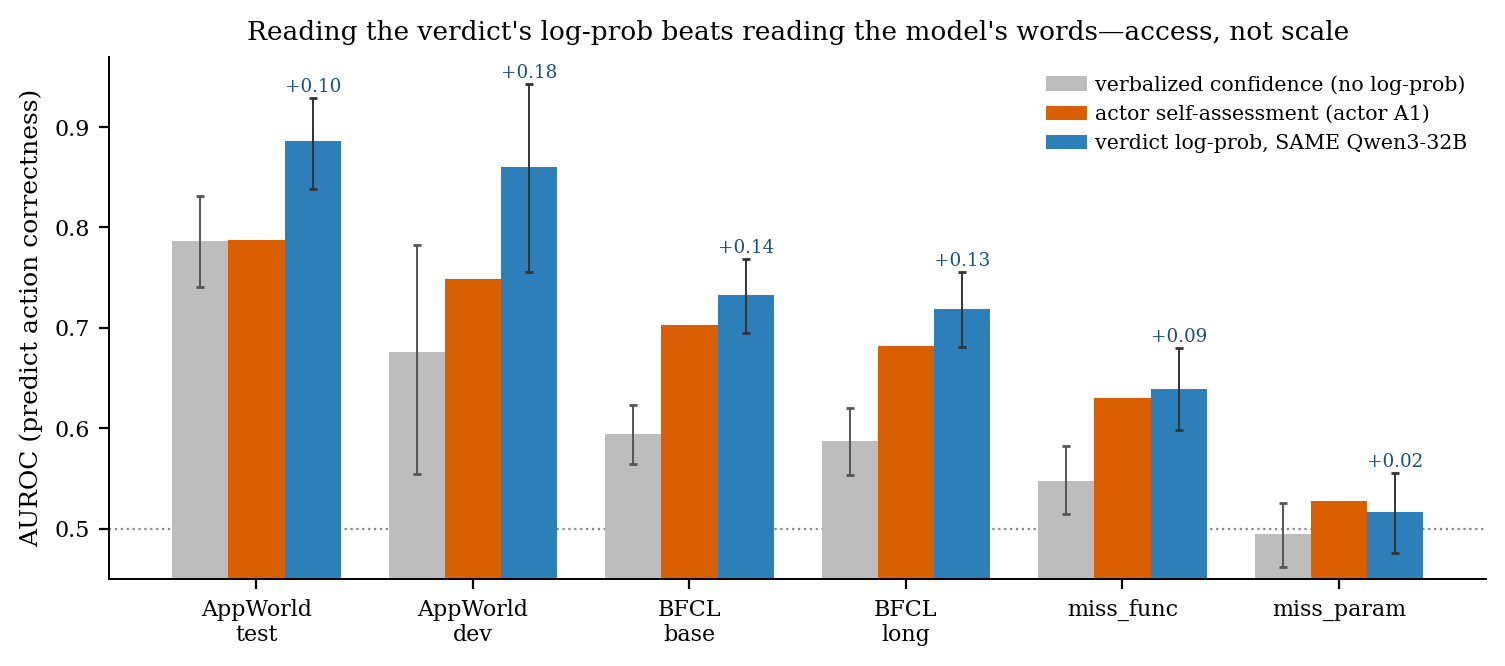}
\caption{\textbf{The gain is access, not scale.} On one Qwen3-32B: the verdict token's log-probability
(blue) vs.\ the same model's \emph{verbalized} confidence (grey) and the actor's own self-assessment
(orange, A1), across six datasets. Capability is held constant between blue and grey.}
\label{fig:ladder}
\end{figure}

\section{Worst-case regret of fixed signal policies}\label{app:regret}
Full breakdown behind the safe-default argument of \S\ref{sec:recipe}. Regret is the AUROC gap between a
fixed policy and the best-achievable signal in each of the six single-call regimes where verdict and TF are
both defined; lower is more robust.

\begin{table}[h]\centering\small
\caption{\textbf{Oriented fusion has no catastrophic regime.} Worst-case regret over the six single-call
regimes where verdict and TF are both defined; lower is more robust.}
\label{tab:regret}
\begin{tabular}{lcc}
\toprule
Fixed policy & worst-case regret (where) & mean regret \\
\midrule
always verdict         & $+0.180$ (miss\_param) & $0.046$ \\
always teacher-forcing & $+0.165$ (BFCL v3)     & $0.087$ \\
\textbf{oriented fusion} & $\mathbf{+0.019}$ (BFCL long) & $\mathbf{0.007}$ \\
\bottomrule
\end{tabular}
\end{table}

\section{Selective prediction: per-dataset results}\label{app:selective}
Per-dataset breakdown behind Mode~A (\S\ref{sec:act}): accept the top-confidence fraction, abstain on the rest.

\begin{table}[h]\centering\small
\caption{\textbf{Ranking by proxy confidence beats random abstention everywhere.} Accept the
top-confidence fraction and abstain on the rest.}
\label{tab:selective}
\setlength{\tabcolsep}{3.5pt}\footnotesize
\begin{tabular}{@{}lcccccc@{}}
\toprule
Dataset & $N$ & base & AUROC & acc@50\% ($\Delta$) & AURC$\downarrow$ & rand \\
\midrule
BFCL (verdict+TF) & 2388 & 0.692 & 0.596 & 0.743 ($+0.051$) & 0.241 & 0.308 \\
xLAM (verdict)    & 412  & 0.367 & 0.713 & 0.524 ($+0.158$) & 0.484 & 0.633 \\
xLAM (needle)     & 725  & 0.542 & 0.890 & 0.843 ($+0.300$) & 0.179 & 0.458 \\
NESTFUL (needle)  & 238  & 0.613 & 0.788 & 0.832 ($+0.218$) & 0.178 & 0.387 \\
tau2 (trajectory) & 149  & 0.772 & 0.679 & 0.867 ($+0.095$) & 0.149 & 0.228 \\
\bottomrule
\end{tabular}
\\[2pt]{\footnotesize acc@50\%: accepted-call accuracy at $50\%$ coverage ($\Delta$ vs.\ base);
AURC$\downarrow$: risk-coverage area (lower is better); rand: random-abstention AURC. \emph{needle} rows
inject one corrupted call, so their base accuracy is measured over that injected-error subset---which is
why the two xLAM rows show different base accuracies (a subset difference, not a measurement one). tau2 is
scored at trajectory level.}
\end{table}

\section{Coverage vs.\ the combiner (battery detail)}\label{app:battery}
Given a battery, the combiner barely matters but orientation is essential (\S\ref{sec:toolkit}). On the one
richly-instrumented, validly-labeled dataset we can test it (BFCL~v3, proxy 32B scorer, $n{=}216$), the
full 16-readout single-pass battery reaches $0.764$ vs.\ $0.706$ for the verdict$+$TF pair
($\Delta{=}{+}0.059$, one-sided $p{=}0.028$; two-sided CI$_{95}$ $[{-}0.001,{+}0.119]$ grazes $0$)---%
\emph{suggestive, not confirmed}. The combiner choice is not distinguishable: a learned logistic scores
$0.733$ against $0.764$ for the orientation-corrected rank-average, with a CI$_{95}$ crossing $0$. A
permutation test (shuffled labels) collapses the learned OOF to $0.50$, so there is no leakage. The larger
tau2 battery result of prior drafts is withdrawn: tau2 lacks a per-call correctness oracle
(\S\ref{sec:limits}).
On the 4-signal xLAM/NESTFUL cells, where few readouts are defined, the full set only ties verdict$+$TF
($\Delta{\le}{+}0.025$). \emph{Per-argument-count (xLAM):} the learned lift over rank-average is
significant on multi-argument calls (CI$_{90}$ $[{+}0.010,{+}0.044]$) but null on single-argument calls
($[{-}0.002,{+}0.032]$)---one value-token readout already dominates there.

\section{Actor self-consistency: detail}\label{app:sc}
\begin{figure}[t]\centering
\includegraphics[width=0.86\linewidth]{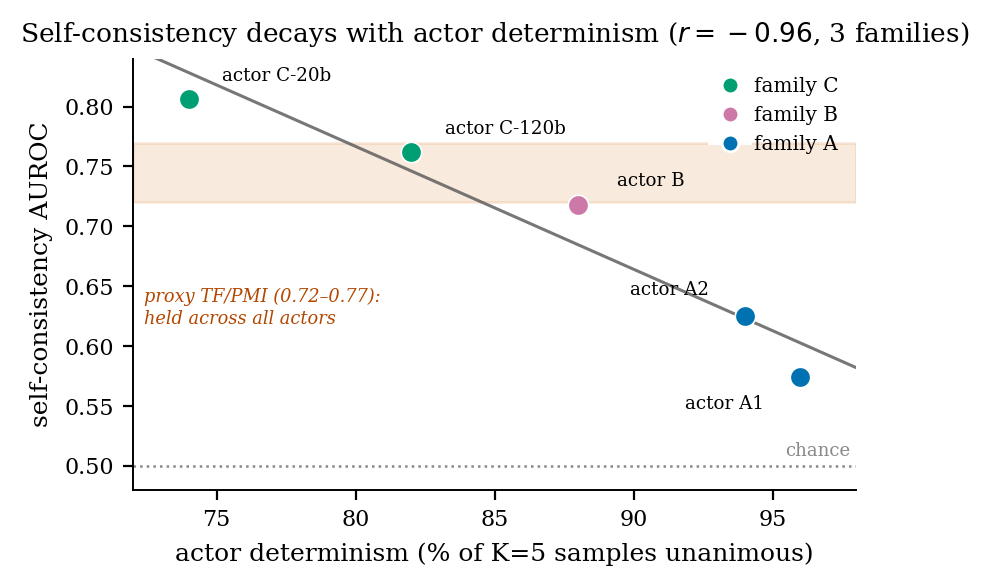}
\caption{\textbf{Self-consistency collapses as the actor becomes deterministic.} SC AUROC vs.\ sample
unanimity (xLAM, $K{=}5$, execution-free; $r{=}{-}0.96$); proxy TF/PMI (shaded) is entropy-invariant.}
\label{fig:scdet}
\end{figure}

Self-consistency (SC) $=$ plurality agreement of $K{=}5$ temp-$0.8$ samples (name$+$canonicalized args),
execution-free; the label is the gold-match correctness of the deployed temp-$0$ call (leave-base-out,
non-circular). SC's AUROC as a correctness predictor \emph{decays monotonically} with how unanimous the
actor's samples are ($r{=}{-}0.96$ across five actors from three families; Figure~\ref{fig:scdet}).
Near-deterministic frontier actors therefore drive SC to chance. On the $\sim\!95\%$ of xLAM problems where
the $K$ samples are unanimous, SC carries \emph{zero} signal by construction, yet the proxy TF signal still
discriminates at $0.72$. \textbf{Code execution artifact.} Cluster-agreement of the $K$ samples \emph{on
the public tests} scores $0.72$--$0.84$ on code\_contests/APPS, but that implicitly runs the public-test
oracle the proxy signal never sees. Made execution-free (agreement on normalized code text), it drops to
$0.52$--$0.64$ on all three code datasets, and on A1 the proxy verdict beats it (LiveCodeBench
$0.851$ vs.\ $0.623$). (The three LiveCodeBench verdict figures measure different comparisons on this same
regime: $0.825$ vs.\ actor self-assessment, $n{=}139$ (\S\ref{sec:beats}); $0.836$ under leave-one-dataset-out
transfer; and $0.851$ here on the SC-comparison subset.)

\section{Actor$\times$scorer robustness grid}\label{app:grid}
The verdict is not an artifact of one actor$\times$scorer pairing (\S\ref{sec:cheap},\ref{sec:beats}).

\begin{table}[t]
\centering\small
\caption{\textbf{The verdict is not one actor$\times$scorer artifact.} Verdict AUROC across actors and
scorers. $P$ $=$ paired-bootstrap exceedance vs.\ that actor's self-assessment; \emph{italic} self $=$ actor
not beaten (self-assesses well).}
\label{tab:grid}
\begin{tabular}{llcccc}
\toprule
Dataset & Actor & Qwen & Gemma & self & $P$(Q/G) \\
\midrule
AppWorld & A1 & \textbf{0.894} & 0.836 & 0.788 & 1.00/.66 \\
AppWorld & A2 & \textbf{0.919} & 0.884 & \emph{0.914} & .57/.14 \\
BFCL base & A1 & 0.733 & 0.728 & 0.703 & .95/.90 \\
BFCL base & A2 & 0.800 & \textbf{0.823} & 0.732 & \textbf{1.00/1.00} \\
BFCL long & A1 & 0.719 & 0.723 & 0.682 & .97/.98 \\
BFCL long & A2 & 0.820 & 0.821 & 0.759 & \textbf{1.00/1.00} \\
\midrule
\multicolumn{6}{l}{\emph{Actors outside family~A (cross-family generalization; \S\ref{sec:beats}):}}\\
BFCL base & B & 0.665 & 0.675 & 0.656 & .68/.81 \\
BFCL long & B & 0.636 & 0.657 & \emph{0.663} & .08/.40 \\
BFCL base & C-120b & 0.739 & 0.735 & \emph{0.740} & .43/.25 \\
BFCL long & C-120b & 0.759 & 0.748 & 0.739 & .94/.74 \\
BFCL base & C-20b & 0.755 & 0.768 & 0.747 & .67/.87 \\
BFCL long & C-20b & 0.753 & 0.766 & \emph{0.765} & .22/.49 \\
\bottomrule
\end{tabular}
\\[2pt]{\footnotesize Bedrock generated-token elicitation; cf.\ matched YES/NO in Table~\ref{tab:cap14}.}
\end{table}

\section{Experimental coverage matrix}\label{app:coverage}
Each result table reports the datasets, actors, and scorers for which its signal is \emph{defined} and its
claim relevant---not a cherry-picked selection. Value-token TF/PMI need a single forced JSON-argument call
(undefined on whole-program code and multi-call renders); the verdict applies to any action.

\begin{table}[t]
\centering\small
\caption{\textbf{Which datasets support which signal, and where each result is used.}}
\label{tab:coverage}
\setlength{\tabcolsep}{3pt}\footnotesize
\begin{tabular}{@{}lcc>{\raggedright\arraybackslash}p{2.7cm}@{}}
\toprule
Dataset & verd. & TF/PMI & used by (actors; scorers) \\
\midrule
AppWorld & \checkmark & --- & grid, hero, Mode~B (A{+}B{+}C; Q,Gm) \\
BFCL base/long & \checkmark & --- & grid, hero, cap14 (A{+}B{+}C; Q,q,Gm) \\
BFCL parallel/v3 & \checkmark & \checkmark & hero, cap14, battery (A; Q,q) \\
BFCL miss\_func/param & \checkmark & \checkmark$^{\dagger}$ & hero, cap14 (A; Q,q) \\
xLAM & \checkmark & \checkmark & hero, cap14, selective, active, SC, \S\ref{sec:cheap} (A{+}B{+}C; Q,q) \\
NESTFUL & \checkmark & \checkmark & hero, cap14, selective, active (A; Q,q) \\
tau2-bench & \S$^{\S}$ & \S$^{\S}$ & scope note only (\S\ref{sec:limits}) \\
LiveCodeBench/cc/APPS & \checkmark & ---$^{\ddagger}$ & hero (\S\ref{sec:beats}), transfer, SC, multi-axis (A; Q) \\
BIRD (SQL) & \checkmark & \checkmark$^{\ddagger}$ & multi-axis verdict$^{\P}$ (A; Q) \\
\bottomrule
\end{tabular}
\\[2pt]\parbox{\columnwidth}{\footnotesize Actors: A$=$A1/A2, $+$B$=$actor~B, $+$C$=$C-120b/C-20b
(generalization only). Scorers: Q$=$Qwen3-32B, q$=$14B, Gm$=$Gemma-27B.
$^{\dagger}$value-mask degenerate on withheld/multi-call args (reported as full-seq or
excluded). $^{\ddagger}$whole-program TF/PMI, not the per-argument value-token TF/PMI of the error-locus study.
$^{\P}$BIRD carries only the multi-axis-verdict contrast in this paper
(App.~\ref{app:signals}); its other two results are a regime-graded \emph{tie} and a corrective
\emph{null}, both consistent with claims we make elsewhere. Auditing works: the proxy verdict reaches
$0.733$ (CI$_{95}$ $[.688,.777]$, $\mathcal{S}_{\text{L}}$ scorer, $n{=}442$; $0.685$ $[.637,.732]$ with the Qwen3-32B
scorer). But the actor self-assesses just as well ($0.715$ $[.665,.760]$; $\Delta{+}0.020$, $P{=}0.81$),
so BIRD is a tie on \emph{beats-self} (\S\ref{sec:beats}). Signal-gated self-review also does not beat
blind or deranged-placebo review ($0.499$ vs.\ $0.510$/$0.515$, $n{=}448$)---the SQL-reasoning instance of
the self-verification boundary of App.~\ref{app:selfreview}.
$^{\S}$tau2-bench lacks a per-call \emph{correctness} oracle (its gold action-checks give an action-salience
label), so it is not used as an auditing benchmark; retained only as a scope note (\S\ref{sec:limits}).}
\end{table}

\section{Capability-axis table (14B vs.\ 32B)}\label{app:cap14}
\begin{figure}[t]
\centering
\includegraphics[width=\linewidth]{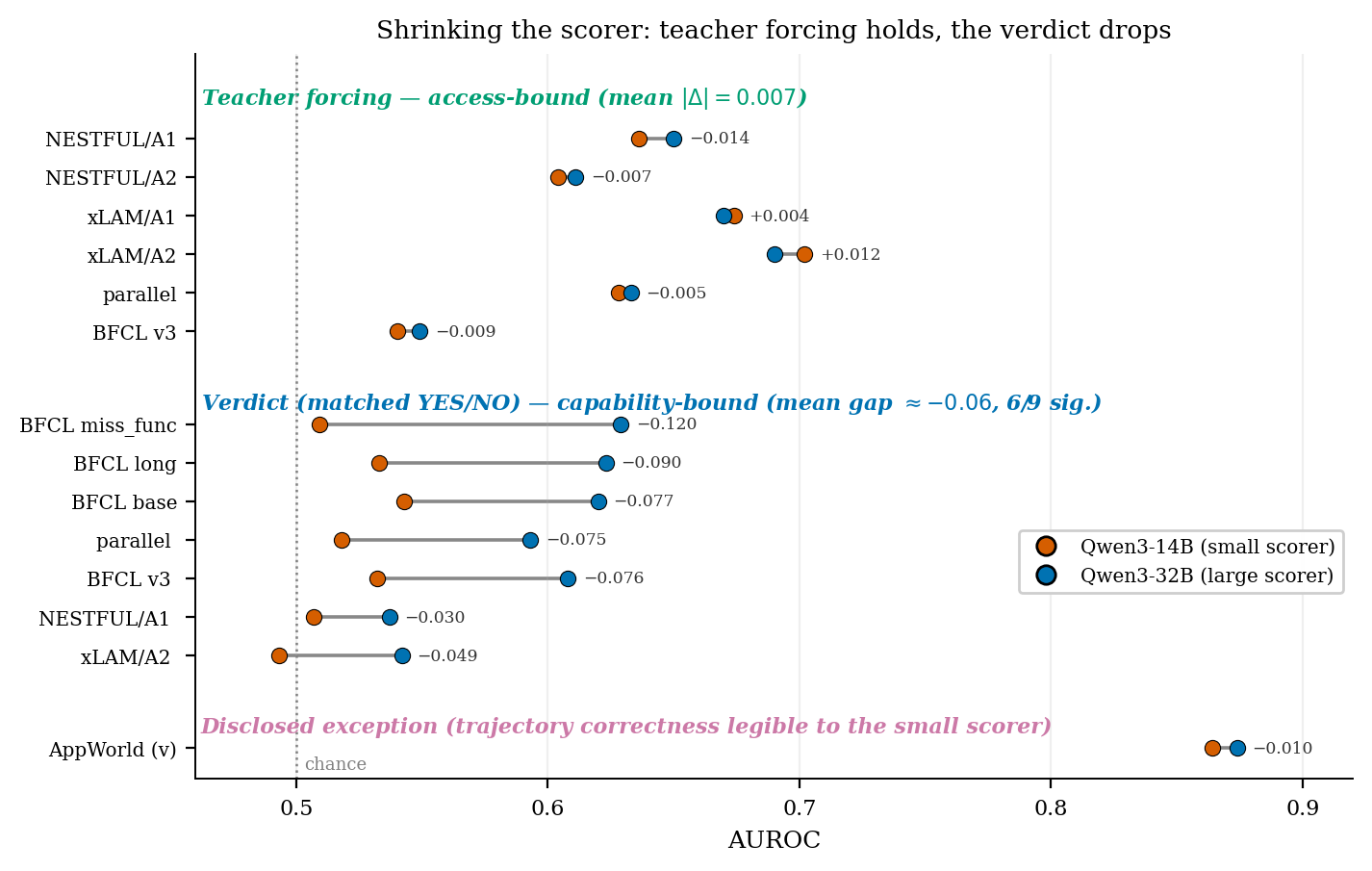}
\caption{\textbf{Teacher forcing is size-invariant; the verdict is not.} Each dumbbell joins 14B
(orange) and 32B (blue) AUROC for one cell, same items and prefill. Top: teacher forcing. Middle: verdict.
Bottom: AppWorld exception. Chance $=0.5$ (dotted).}
\label{fig:split}
\end{figure}

Representative matched-mechanism cells behind the capability split (Figure~\ref{fig:split}); see
\S\ref{sec:cheap} for the full $6/6$ TF and $6/9$ verdict tallies.

\begin{table}[t]
\centering\small
\caption{\textbf{14B matches 32B on teacher forcing, trails on request-PMI and the verdict.} Qwen3-14B
vs.\ 32B, matched items per signal. $\Delta{=}$14B$-$32B; verdict uses matched YES/NO teacher-forcing. Full
tally in \S\ref{sec:cheap}.}
\label{tab:cap14}
\begin{tabular}{llccc}
\toprule
Dataset & Signal & 14B & 32B & $\Delta$ \\
\midrule
\multicolumn{5}{l}{\emph{Teacher forcing---access-bound ($|\Delta|{\le}0.015$, TOST-equivalent)}}\\
NESTFUL/A1 & TF$_{\text{val}}$ & 0.636 & 0.650 & $-.014$ \\
NESTFUL/A2 & TF$_{\text{val}}$ & 0.604 & 0.611 & $-.006$ \\
xLAM/A1 & TF$_{\text{val}}$ & 0.674 & 0.670 & $+.004$ \\
xLAM/A2 & TF$_{\text{val}}$ & 0.702 & 0.690 & $+.011$ \\
parallel/A1 & TF$_{\text{val}}$ & 0.628 & 0.633 & $-.005$ \\
BFCL v3/A1 & TF$_{\text{val}}$ & 0.540 & 0.549 & $-.009$ \\
\midrule
\multicolumn{5}{l}{\emph{Request-PMI---scale-sensitive (contrasts two prefills)}}\\
xLAM/A2 & PMI$_{\text{min}}$ & 0.703 & 0.751 & $-.048$ \\
BFCL v3 & PMI$_{\text{min}}$ & 0.650 & 0.689 & $-.039$ \\
\midrule
\multicolumn{5}{l}{\emph{Discriminative---capability-bound (verdict, matched mechanism)}}\\
BFCL base & verdict & 0.543 & 0.620 & $-.077$ \\
BFCL miss\_func & verdict & 0.509 & 0.629 & $-.120$ \\
xLAM/A2 & verdict & 0.493 & 0.542 & $-.049$ \\
AppWorld & verdict & 0.864 & 0.874 & $-.010$ \\
\bottomrule
\end{tabular}
\end{table}

\section{Sidecar latency measurement}\label{app:latency}
Per-call scoring latency behind the cost argument of \S\ref{sec:recipe}, measured on the deployed sidecar
at realistic full-context length. A scored call is a single prefill of ${\approx}160$\,ms, overlapped with
tool I/O in Mode~A.

\begin{table}[t]\centering\small
\caption{\textbf{A scored call is a single prefill of ${\approx}160$\,ms.} Measured sidecar latency at
realistic full-context length (${\approx}1.5$k tokens; Qwen3-14B fp8, one L40S GPU).}
\label{tab:latency}
\begin{tabular}{@{}lccc@{}}
\toprule
Readout & $p50$ & $p90$ & $p99$ \\
\midrule
Teacher forcing & $159$\,ms & $166$\,ms & $179$\,ms \\
Verdict         & $123$\,ms & $132$\,ms & $137$\,ms \\
\bottomrule
\end{tabular}
\\[2pt]{\footnotesize Concurrency~1; $40$ timed calls after warm-up.}
\end{table}

\section{Feeding the signal back to the actor (self-review detail)}\label{app:selfreview}
Beyond selective escalation (Mode A) and confidence feedback (Mode B), we tested \emph{handing the actor its own
proxy confidence to self-correct}. The protocol keeps the actor black-box: it completes its trajectory;
the Qwen3-32B auditor scores every call; the actor is then shown its calls annotated with per-call signals
and re-emits a corrected sequence. Arms differ only in the annotation: \texttt{blind} (none),
\texttt{verdict}/\texttt{tf}/\texttt{fusion}/\texttt{all}, and \texttt{placebo}. The placebo shows the same
numbers deranged onto the wrong calls---identical flag budget, wrong positions---which makes it the decisive
control: it separates localized information from a mere look-again nudge.

The boundary is whether the actor can \emph{self-verify the call at review time} (Table~\ref{tab:active}).
Take single-shot xLAM first, where correctness is legible only from the schema. There, showing the signal
beats unaided review on the A1 reviewer (Holm $p{=}0.004$ vs.\ blind), and beats the deranged placebo on
both reviewers (A1 $p{=}0.013$, A2 $p{=}0.006$). The gain is therefore localized information rather
than a look-again nudge. On A2 the vs-blind margin is directional but not significant after correction.
NESTFUL (recompute) and live BFCL-v4 (re-read real tool output) go the other way: unaided review already
captures the gain, and no signal arm beats blind or placebo. These negatives share one cause, and it is the
boundary itself---when the actor can cheaply self-verify the call from the real result, a proxy number adds
little beyond unaided reflection. The same boundary separates our two inline benchmarks below.

\textbf{Mode~B in full: the readouts in a live loop.} Detail behind \S\ref{sec:act}'s Mode~B---appending
each step's confidence to its real tool result, so the actor sees it in context on every observation and may
adjust its next action. We compare the readouts of \S\ref{sec:toolkit}
against three controls: an unannotated run, a \emph{vigilance} arm that announces the audit but shows no
number, and a \emph{random} arm that shows an independent $U(0,1)$ draw in place of the real confidence
(Table~\ref{tab:inline}).

On InterCode \textsc{nl2bash}~\citep{yang2023intercode} the readouts order as \S\ref{sec:toolkit} predicts. All arms cover the same 193 usable tasks (200 minus 7 whose gold command fails
its own grader) and every contrast is paired. Announcing an audit without showing a number does nothing
($0.389$, $-0.010$, $p{=}0.87$), so any effect requires the number itself. The discriminative verdict alone
does not separate ($0.430$, $+0.031$, $p{=}0.36$). The generative likelihoods do: request-PMI reaches $0.472$
($+0.073$, 25/11, $p{=}0.029$) and teacher forcing $0.497$ ($+0.098$, 25/6, $p{=}0.001$). Strongest is the
oriented fusion at $0.518$ ($+0.119$, 29 rescues vs.\ 6 regressions, exact $p{=}10^{-4}$, bootstrap
95\% CI $[{+}0.062,{+}0.176]$). The generative-over-discriminative ordering and the fusion's low-regret
behaviour were both established offline as ranking properties (\S\ref{sec:toolkit}, \S\ref{sec:cheap}); here
they reappear in a live loop. Read against the random arm below, only the generative readouts and their
fusion clear its $0.440$---the verdict arm ($0.430$) does not reach it. The separation that survives a noise
control is therefore specifically the generative one.

On \textsc{nl2bash} the random arm rules out the trivial explanation that any number in the loop makes the actor more careful:
an independent $U(0,1)$ draw does not significantly beat the unannotated control ($0.440$, $+0.041$, 25/17,
$p{=}0.28$), and fusion beats it by $+0.078$ (19/4, $p{=}0.003$, CI $[{+}0.031,{+}0.124]$). A meaningless
scalar is therefore not sufficient: the annotation's content matters relative to noise. We state the claim at
that level deliberately. The real and random streams differ in more than one respect---in where their low
values fall, and in how they are scaled. (Fusion occupies a narrow band, mean $0.636$, sd $0.144$, while the
uniform draw has twice the spread and dips below $0.10$ on $9\%$ of steps.) The comparison therefore does not
by itself identify which property drives the effect, so we report the end-to-end result and leave the causal
decomposition to future work.

AppWorld \texttt{test\_challenge} (417 tasks, 139 scenarios) shows the opposite pattern, and the contrast is
the point. Inline fusion gains $+0.137$ over the unannotated control there ($100$/$43$, $p{<}10^{-4}$). But so
does the random arm, by $+0.132$ ($95$/$40$, $p{<}10^{-4}$), and the two are indistinguishable
($+0.005$, $46$/$44$, $p{=}0.92$; cluster-robust 95\% CI $[{-}0.043,{+}0.053]$, and $+0.008$, $p{=}0.81$, on
the 390 tasks where our step cap forced neither label). On that benchmark any scalar suffices: we detect no
contribution from the readout's content. Reporting only the vs-control number would have made AppWorld look
like a second confirmation; the random arm shows it is not one.

What distinguishes \textsc{nl2bash} is the shape of its failures: its answers are md5 sums, counts and
file listings, so a subtly wrong pipeline returns a well-formed, plausible, wrong value. Classifying every
executed step of the benchmark's published reference trajectories ($n{=}1176$), \textbf{$50\%$ of wrong steps
are \emph{silent}}---clean, non-empty output with no error text---and $48\%$ of episodes contain at least
one such step (a proxy for our actor's error profile rather than a direct measurement of it). There is no
traceback for a reviewer to restate, so an external judge adds information the actor cannot obtain from its
own observation. (Our unannotated control scores $0.399$, below the $0.485$ the benchmark reports for its own published reference actor.
Two harness choices account for the gap. The first is an 8-turn cap per task. The second is rewriting
leading-\texttt{cd} compound commands as subshells, to work around a path-handling bug in the environment
that prevents the working directory from persisting. Both apply identically to every arm, so they depress
absolute success uniformly and leave the paired contrasts unaffected.)

The two benchmarks also differ in the signal's measured validity, in the same direction. On AppWorld the
\emph{displayed} inline confidence has no task-level validity on this split (AUROC $0.388$ mean-pooled,
$0.474$ min-pooled---at or below chance, and below a trajectory-length baseline at $0.608$), against $0.64$
on \textsc{nl2bash}. This is a different quantity from the AppWorld verdict of Table~\ref{tab:hero}
($0.894$): that is a per-call verdict AUROC on the standard split, whereas this is the task-level pooled
validity of the displayed fusion on \texttt{test\_challenge} under the revised scorer prompt. The per-call
readout retains its ranking power; what fails is its aggregation to a task-level target on this split. The
reason is structural: AppWorld steps execute Python whose failures mostly \emph{raise}, so the diagnostic
information is already in the observation the actor reads. A scorer-prompt revision that removed a measured
position artifact there (confidence--step-index correlation $+0.60\rightarrow+0.15$, reaching parity between
mandatory setup steps and substantive actions) succeeded mechanically yet bought \emph{zero} predictive
power. That locates the failure in the target rather than the prompt. The operational reading is this: the
readout's \emph{content} earns its cost only where the environment does not already surface its own errors.
An inline number may still raise accuracy elsewhere, as it does on AppWorld, but then any scalar would do.
A random-value arm is what tells an operator which case they are in.

Taken together, Mode~B raises end-to-end accuracy on both live-execution benchmarks ($+0.119$ and
$+0.137$, both $p{\le}10^{-4}$). But only on \textsc{nl2bash} does that gain depend on \emph{what} the number
says. The dividing line is whether the environment already surfaces its own errors, and it coincides with
the signal's measured validity ($0.64$ vs.\ chance). One evaluation lesson generalizes: for such loops an
unannotated control is not sufficient, and a random-value arm is the minimum. ``Beats no annotation'' and
``beats noise'' are different claims, and on one of our two benchmarks they come apart.

\begin{table}[t]\centering\small
\caption{\textbf{Inline readout comparison, and why a random-value arm is required.} A2 actor; each
readout appended to the real tool result. Contrasts are paired against the unannotated control;
\textsc{nl2bash} arms all cover the same 193 tasks, AppWorld all 417. \emph{Vigilance} announces the audit
but shows no number; \emph{random} shows an independent $U(0,1)$ draw. The two benchmarks diverge on the
random arm: on \textsc{nl2bash} it is flat and fusion beats it, so the annotation's content matters; on
AppWorld it matches fusion, so there any scalar suffices. Against the unannotated control alone both
benchmarks would have looked alike.}
\label{tab:inline}
\setlength{\tabcolsep}{4pt}\footnotesize
\begin{tabular}{@{}lccccc@{}}
\toprule
 & \multicolumn{2}{c}{\textsc{nl2bash}} & & \multicolumn{2}{c}{AppWorld \texttt{test\_challenge}} \\
\cmidrule(r){2-3}\cmidrule(l){5-6}
Arm & success & vs.\ control & & success & vs.\ control \\
\midrule
control (no number)     & $0.399$ & ---                       & & $0.595$ & --- \\
vigilance (no number)   & $0.389$ & $-0.010$ n.s.             & & ---     & --- \\
random $U(0,1)$         & $0.440$ & $+0.041$ n.s.             & & $0.727$ & $+0.132$ ($p{<}10^{-4}$) \\
verdict only            & $0.430$ & $+0.031$ n.s.             & & ---     & --- \\
request-PMI             & $0.472$ & $+0.073$ ($p{=}.029$)     & & ---     & --- \\
teacher forcing         & $0.497$ & $+0.098$ ($p{=}.001$)     & & ---     & --- \\
\textbf{fusion}         & $\mathbf{0.518}$ & $\mathbf{+0.119}$ ($p{=}10^{-4}$) & & $\mathbf{0.731}$ & $\mathbf{+0.137}$ ($p{<}10^{-4}$) \\
\midrule
\textbf{fusion vs.\ random} & \multicolumn{2}{c}{$\mathbf{+0.078}$, 19/4, $\mathbf{p{=}.003}$} & & \multicolumn{2}{c}{$+0.005$, 46/44, $p{=}.92$} \\
signal validity (AUROC) & \multicolumn{2}{c}{$0.64$ (task-level)} & & \multicolumn{2}{c}{$0.39$--$0.47$ (chance)} \\
\bottomrule
\end{tabular}
\end{table}

\begin{table}[t]\centering\small
\caption{\textbf{Feedback helps only when the actor cannot self-verify.} Signal-guided self-review vs.\
deranged-placebo, \texttt{all}-signal arm. Self-verif.\ $=$ can the actor check the call at
review time. Bottom: repair rate at a one-call audit budget, signal- vs.\ random-gated.}
\label{tab:active}
\setlength{\tabcolsep}{3pt}\footnotesize
\begin{tabular}{@{}l>{\raggedright\arraybackslash}p{1.25cm}cc>{\raggedright\arraybackslash}p{1.95cm}@{}}
\toprule
Dataset & Self-verif.? & \texttt{blind} & \texttt{all} & \texttt{all} vs.\ blind\,/\,placebo (Holm $p$) \\
\midrule
xLAM, A1 ($n{=}196$)  & no (schema)   & $0.066$ & $\mathbf{0.148}$ & $0.004$/$0.013$ \\
xLAM, A2 ($n{=}196$)    & no (schema)   & $0.097$ & $0.138$ & $0.46$/$0.006$ \\
NESTFUL ($n{=}184$)       & yes (recomp.)    & $0.844$ & $0.853$ & n.s.\ / n.s. \\
BFCL-v4 ($n{=}1499$)      & yes (real res.) & $0.892$ & $0.892$ & n.s.\ (all$\approx$) \\
\midrule
\multicolumn{5}{@{}l}{\emph{Selective review at fixed budget $k{=}1$ call (signal- vs.\ random-gated):}}\\
xLAM ($n{=}192$)     & --- & \multicolumn{2}{c}{sig $0.92$ / rand $0.42$} & $p{<}10^{-4}$ (95/1) \\
NESTFUL ($n{=}69$)   & --- & \multicolumn{2}{c}{sig $0.65$ / rand $0.40$} & $p{=}0.008$ (27/10) \\
\bottomrule
\end{tabular}
\end{table}

\section{Signals beyond the single-pass family, and the judge-free anchor}\label{app:signals}

\begin{figure}[t]\centering
\includegraphics[width=\linewidth]{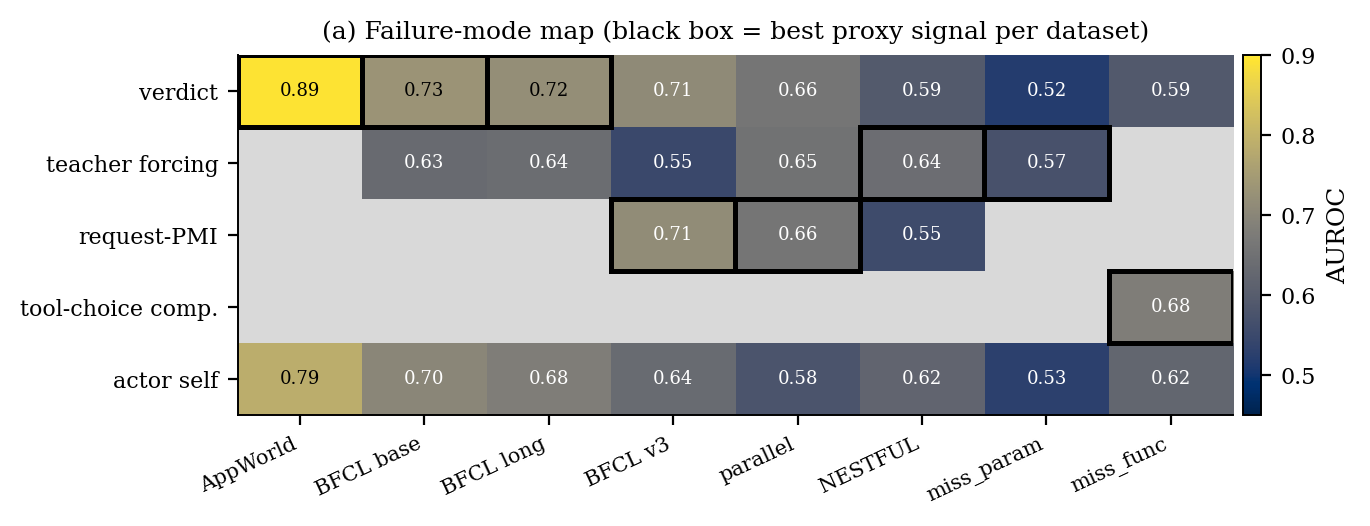}
\caption{\textbf{Failure-mode map (same data as Table~\ref{tab:hero}).} Signal $\times$ dataset AUROC;
black box $=$ best proxy signal per dataset; grey $=$ signal undefined for that action type.}
\label{fig:heatmap}
\end{figure}

\begin{figure}[t]\centering
\includegraphics[width=0.72\linewidth]{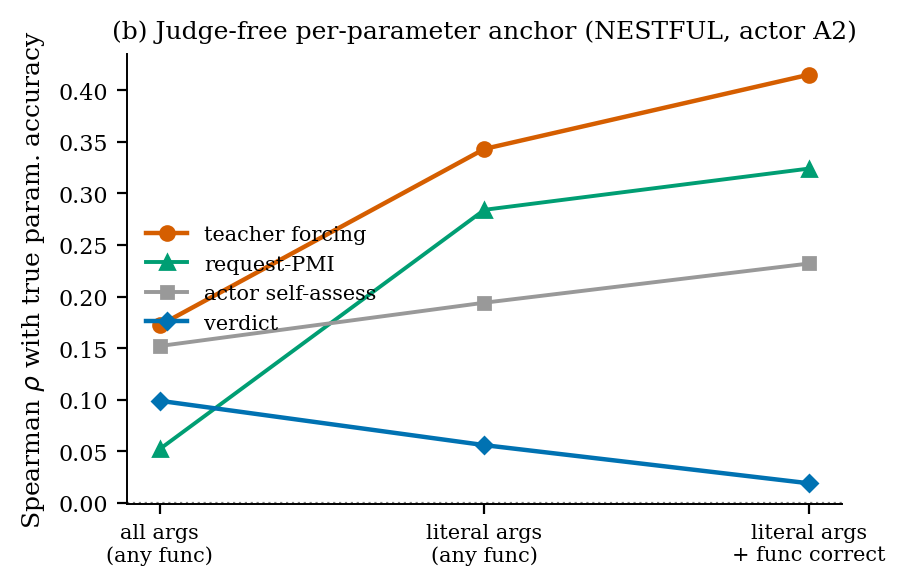}
\caption{\textbf{Teacher forcing tracks argument-value accuracy; the verdict does not.} Spearman $\rho$
vs.\ deterministic per-parameter accuracy (no LLM judge) as the target sharpens to literal argument values
(NESTFUL, A2). TF rises to $\rho{=}0.41$; verdict $0.02$.}
\label{fig:perparam}
\end{figure}

Because the auditor is open it admits four readouts a black-box pipeline cannot compute
(Table~\ref{tab:newsignals}). Ensemble disagreement---$-|q_{\text{Qwen}}{-}q_{\text{Gemma}}|$
between different-lineage scorers---is a strong standalone signal ($0.70$--$0.80$) and near-free.
Semantic entropy of regenerations ($0.68$--$0.71$) tracks self-consistency but does not exceed
single-pass signals. Schema-satisfaction is null on frontier actors ($0.50$--$0.51$), which emit
schema-valid calls $\sim$100\% of the time. Two verdict elaborations mostly do not beat the one-token
verdict. A debiased verdict (symmetric double-ask) does not improve on tool calls (BFCL v3 $0.670$
vs.\ $0.711$). A multi-axis verdict helps only on long code: on LiveCodeBench it gains
($0.818$ vs.\ plain $0.783$), whereas on BIRD text-to-SQL it \emph{trails} ($0.648$ vs.\ $0.670$), both
same-file comparisons.
So the single-token verdict is a strong, cheap default, and elaborations pay off only when the extra axes
carry independent information.

\begin{table}[t]
\centering\small
\caption{\textbf{Extra readouts an open scorer enables, and verdict variants.}}
\label{tab:newsignals}
\setlength{\tabcolsep}{4pt}\footnotesize
\begin{tabular}{@{}lccl@{}}
\toprule
Signal & AUROC & type & note \\
\midrule
Ensemble disagreement & 0.70--0.80 & epistemic & orthogonal, near-free \\
Self-consistency ($k{=}5$) & 0.71--0.75 & epistemic & good on easy calls \\
Semantic entropy & 0.68--0.71 & epistemic & no gain over above \\
Schema-satisfaction & 0.50--0.51 & determ.\ & null (frontier actors) \\
\midrule
\multicolumn{4}{l}{\emph{verdict variants (vs.\ plain verdict, same rows)}}\\
Debiased verdict & 0.51--0.67 & discrim.\ & no gain on tool calls \\
Multi-axis verdict & 0.65--0.82 & discrim.\ & helps on long code only \\
\bottomrule
\end{tabular}
\end{table}

\end{document}